\documentclass[journal]{IEEEtran}

\usepackage[american]{babel}
\usepackage{tabularx,booktabs}
\usepackage{multirow}
\usepackage{color}
\usepackage[breaklinks=true,colorlinks,bookmarks=false]{hyperref}
\usepackage{amsmath,amssymb}
\usepackage[bold]{hhtensor}
\usepackage{graphicx}
\usepackage{amsmath}
\usepackage{algorithmic}
\usepackage[ruled,linesnumbered]{algorithm2e}

\usepackage{array} 
\usepackage{cite}
\usepackage{xspace}
\usepackage{hyperref}  
\usepackage{url}

\ifCLASSOPTIONcompsoc 
  \usepackage[caption=false,font=normalsize,labelfont=sf,textfont=sf]{subfig}
\else
  \usepackage[caption=false,font=footnotesize]{subfig}
\fi

\makeatletter
\let\NAT@parse\undefined
\makeatother

\newcommand{\squishlist}{
 \begin{list}{$\bullet$}
  { \setlength{\itemsep}{0pt}
     \setlength{\parsep}{1pt}
     \setlength{\topsep}{1pt}
     \setlength{\partopsep}{0pt}
     \setlength{\leftmargin}{1.5em}
     \setlength{\labelwidth}{1em}
     \setlength{\labelsep}{0.5em} } }
\newcommand{\squishend}{
  \end{list}  }

\makeatletter
\DeclareRobustCommand\onedot{\futurelet\@let@token\@onedot}
\def\@onedot{\ifx\@let@token.\else.\null\fi\xspace}

\def\eg{\emph{e.g}\onedot} 
\def\ie{\emph{i.e}\onedot}

\begin{document}

\title{APANet: Adaptive Prototypes Alignment Network for Few-Shot Semantic Segmentation}

\author{Jiacheng Chen\textsuperscript{*}, Bin-Bin Gao\textsuperscript{*}, Zongqing Lu, Jing-Hao Xue,~\IEEEmembership{Senior Member,~IEEE}, Chengjie Wang \\ and Qingmin Liao,~\IEEEmembership{Senior Member,~IEEE}%
\thanks{J.~Chen, Z.~Lu and Q.~Liao are with Shenzhen International Graduate School, Tsinghua University, Shenzhen 518055, China (e-mail: cjc19@mails.tsinghua.edu.cn, luzq@sz.tsinghua.edu.cn, liaoqm@tsinghua.edu.cn)}
\thanks{B.-B.~Gao and C.~Wang are with Tencent YouTu Lab, Shenzhen 518057, China (e-mail: gaobb@lamda.nju.edu.cn, jasoncjwang@tencent.com)}%
\thanks{J.-H.~Xue is with Department of Statistical Science, University College London, London WC1E 6BT, UK (e-mail: jinghao.xue@ucl.ac.uk)}%
\thanks{\textsuperscript{*}First two authors contributed equally. This work was done when J.~Chen was an intern at Tencent YouTu Lab.  This
research was supported in part by Tencent and the Special Foundation for the Development of Strategic Emerging Industries of Shenzhen (No. JCYJ20170817161056260). Bin-Bin Gao and Zongqing Lu are the Corresponding authors.}
}

\markboth{Accepted by IEEE Trans. on Multimedia}{Chen \MakeLowercase{\textit{et al.}}}

\maketitle

\begin{abstract}
Few-shot semantic segmentation aims to segment novel-class objects in a given query image with only a few labeled support images. Most advanced solutions exploit a metric learning framework that performs segmentation through matching each query feature to a learned class-specific prototype. However, this framework suffers from biased classification due to incomplete feature comparisons. To address this issue, we present an adaptive prototype representation by introducing class-specific and class-agnostic prototypes and thus construct complete sample pairs for learning semantic alignment with query features. The complementary features learning manner effectively enriches feature comparison and helps yield an unbiased segmentation model in the few-shot setting. It is implemented with a two-branch end-to-end network (\ie, a class-specific branch and a class-agnostic branch), which generates prototypes and then combines query features to perform comparisons. In addition, the proposed class-agnostic branch is simple yet effective. In practice, it can adaptively generate multiple class-agnostic prototypes for query images and learn feature alignment in a self-contrastive manner. Extensive experiments on PASCAL-5$^i$ and COCO-20$^i$ demonstrate the superiority of our method. At no expense of inference efficiency, our model achieves state-of-the-art results in both 1-shot and 5-shot settings for semantic segmentation.
\end{abstract}

\IEEEpeerreviewmaketitle

\section{Introduction}
\IEEEPARstart{F}ew-shot Semantic Segmentation~(FSS)~\cite{OSLSM, panet, canet} has attracted much attention because it makes pixel-level semantic predictions possible for novel classes in the test images~(\emph{query}) with only a few~(\eg, 1 or 5) labeled images~(\emph{support}). Its learning paradigm aims to quickly adapt the model to the novel classes that have never been seen during training and use only a few labeled images during testing. This greatly reduces the requirement for collecting and annotating a large-scale dataset. 

\begin{figure}[t]
\centering
\centerline{\includegraphics[width=1.0\linewidth]{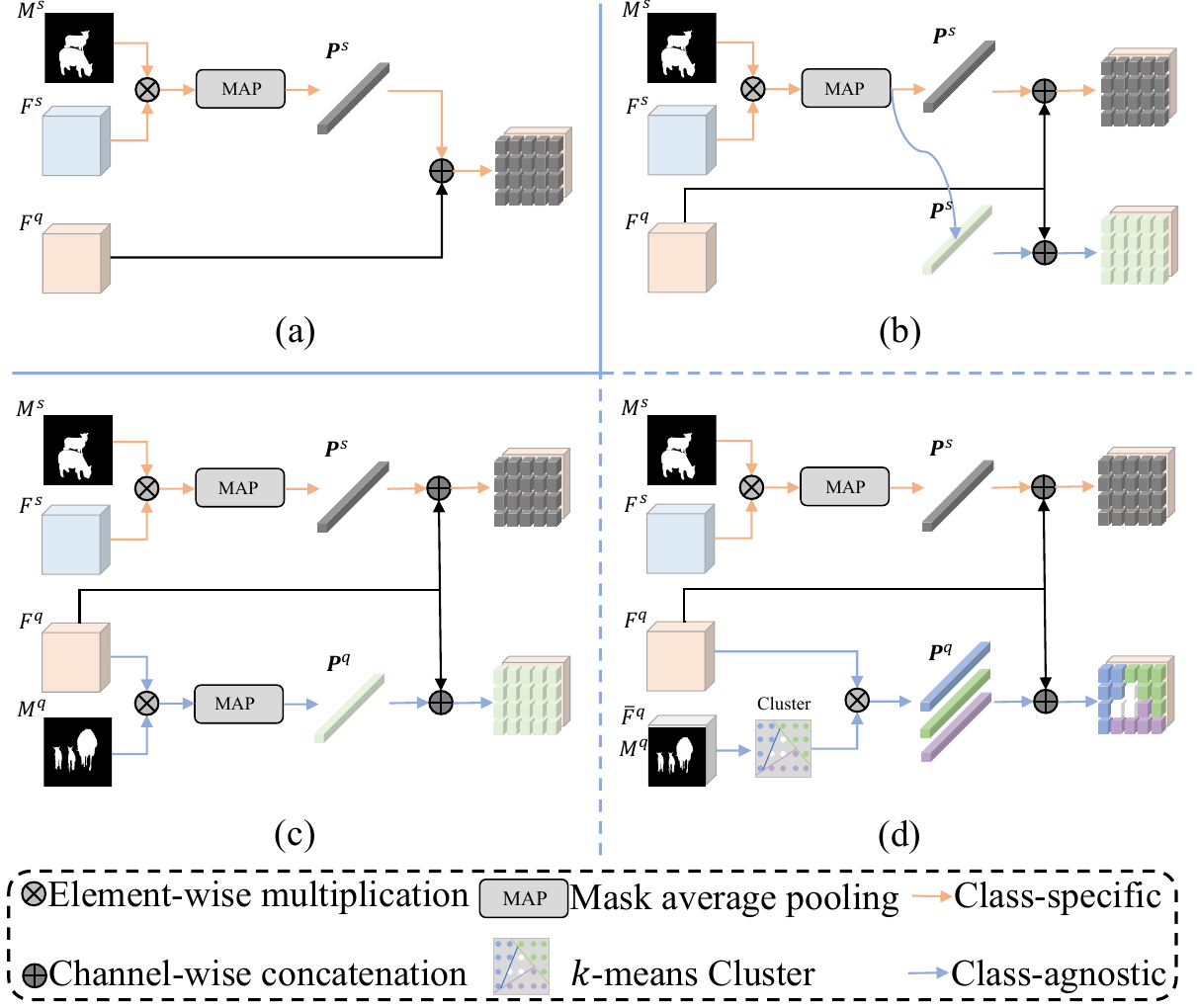}}
\caption{Comparison between \textbf{(a)} common few-shot learning approach and \textbf{(d)} our adaptive prototypes alignment method. The common few-shot learning approach compares features between the class-specific prototype and the query features. In contrast, the proposed method implements comparison not only between the class-specific prototype and the query features but also between multiple class-agnostic prototypes and the query features. Our method adopts a two-branch architecture, which includes a class-specific branch and a class-agnostic branch. The learning process of the class-specific branch is similar to the common few-shot learning. The class-agnostic branch is guided by a self-contrastive manner and its multiple class-agnostic prototypes are adaptively generated by using the $k$-means algorithm. The~\textbf{(b)} and~\textbf{(c)} are two special cases of our proposal in \textbf{(d)}. In~\textbf{(b)} and~\textbf{(c)}, the number of clusters is set to 1, and the difference between them is that the class-agnostic prototype comes from the support image or the query one.}
\label{fig:compareFeature}
\end{figure}

Most FSS methods~\cite{panet, canet, pfenet, fwb, PL} are built on metric learning for its simplicity and effectiveness, by learning to compare the query with a few support images. First, a shared convolution network is used to concurrently extract deep representations of support and query images. Then, these support features and their masks are encoded to a single vector, which forms a foreground class-specific~\emph{prototype}. Finally, a pixel-level comparison is densely conducted between the class-specific prototype and each location of query features, to determine whether or not they are from the same category. The pixel-level feature comparison may be explicit,~\eg, by using the cosine similarity~\cite{PL,panet}, as well as implicit,~\eg, via a relation network~\cite{sung2018learning}. This learning process can be illustrated as in Fig.~\ref{fig:compareFeature}(a). Following this framework, some works try to generate a finer class-specific prototype with support features~\cite{pmms, PGNet, ppnet}. However, these methods suffer from limited generalization ability due to incomplete feature comparison only between the foreground prototype~(generating from support features) and query features.

\begin{figure}[t]
\centering
\centerline{\includegraphics[width=1.0\linewidth]{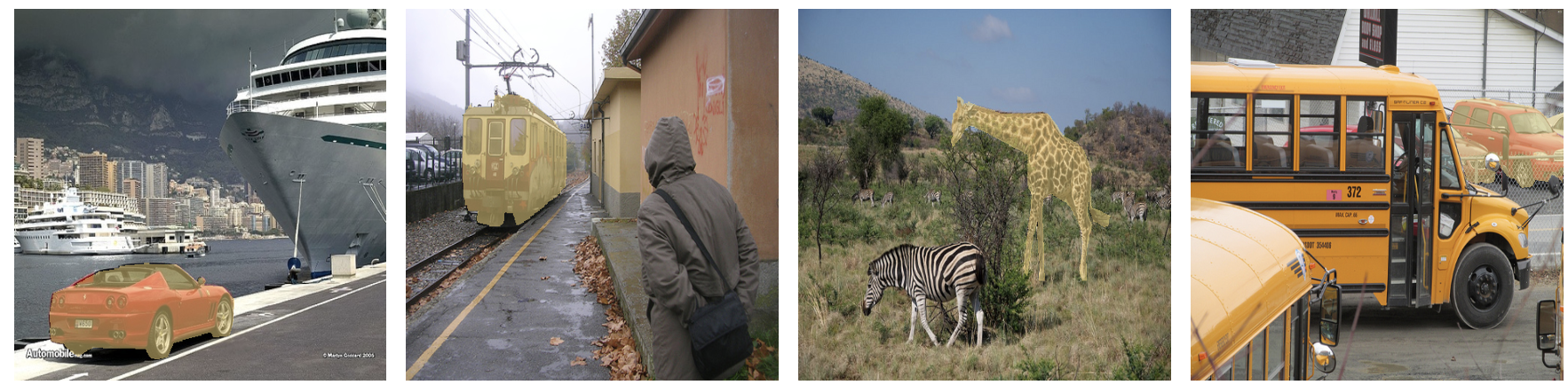}}
\caption{Selected training images on PASCAL-5$^i$ (first two columns) and COCO-20$^i$ (last two columns). The base-class objects, such as car, train and giraffe, are covered by the yellow masks (manual annotation). During training, they are treated as foreground, and others are viewed as background. We easily find that some classes, such as ship, person, zebra and bus, appear in the background of these training images, while they can actually be novel classes that need to be recognized as foreground at the testing stage. This conflict between the training and testing phases will bring a systematic bias, thereby limiting the generalization performance of the few-shot learning model.}
\label{fig:Novel-base}
\end{figure}

\begin{figure}[t]
\centering
\centerline{\includegraphics[width=1.02\linewidth]{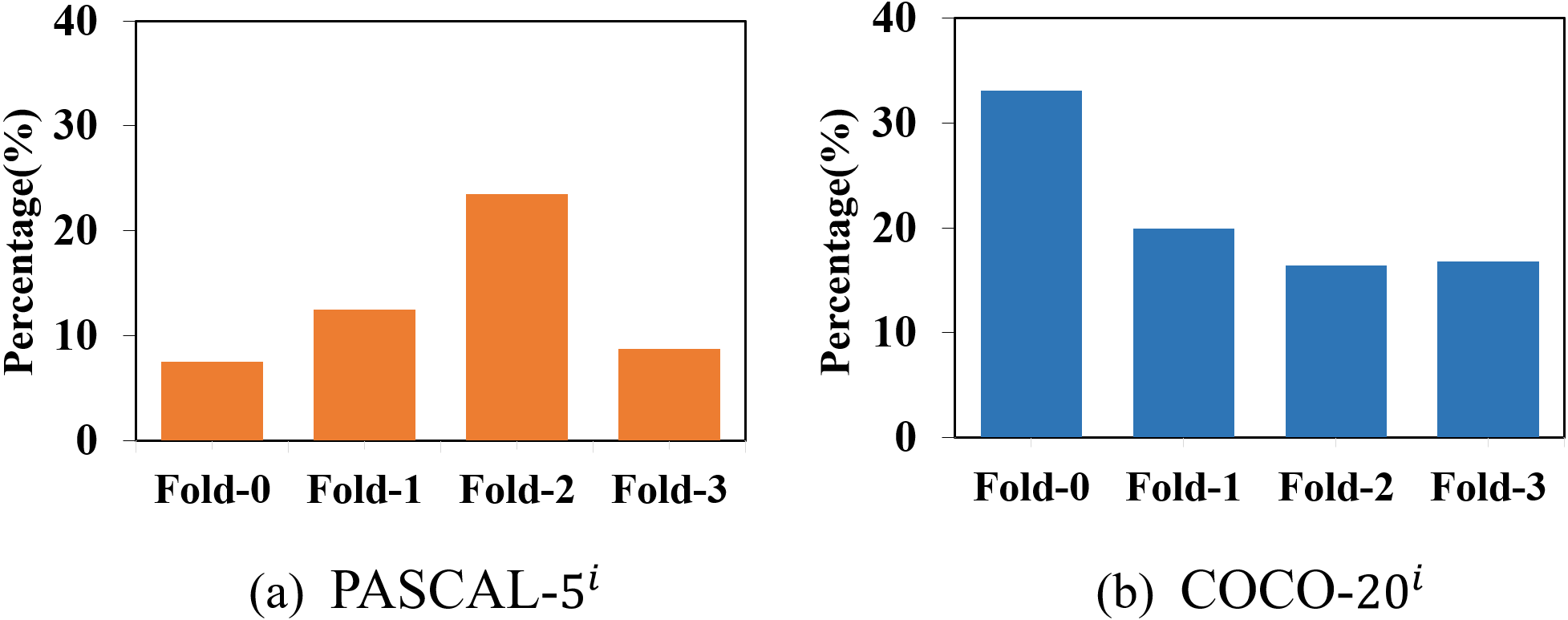}}
\caption{The percentage of potential novel-class objects in the training images. It can be observed that there is a high percentage of novel-class objects in each base fold.} 
\label{fig:STASTICS}
\end{figure}

Why is the generalization limited in most existing works? As shown in Fig.~\ref{fig:compareFeature}(a), the feature comparison is conducted between the foreground prototype and query features, and the whole background features in the query image are treated as negative samples during training. This will lead to an issue with FSS, because it is possible that some novel-class objects in the test set are presented in the training images but treated as background during training. As illustrated in Fig.~\ref{fig:Novel-base}, we can see that some novel-class objects, such as ship, person, zebra and bus, appear in the training images and they are usually viewed as background by many existing few-shot segmentation methods. Specifically, we compute the proportion of novel classes in the training images for each fold on the PASCAL-5$^{i}$ and COCO-20$^i$ datasets according to the original multi-labels of each image. It can be observed that there is a high percentage of novel-class objects in each base fold, as shown in Fig.~\ref{fig:STASTICS}. For example, 23.5\% of novel-class objects remain hidden in the training images of Fold-2 on PASCAL-5$^i$. Similar phenomena can be also observed on COCO-20$^i$. Therefore, it is unsurprising that, during testing, existing methods tend to incorrectly recognize novel-class objects as background, even when the (foreground) novel-class prototype is provided. This will bring very serious system biases that these novel-class objects are difficult to be correctly predicted as foreground at the testing stage, because the system has remembered these potential objects (novel class) as background in the period of training. The biased classification issue also presents in zero-shot object detection (ZOD)~\cite{ZOD}. They iteratively assign background proposals to latent classes with semantic information extracted from a pre-trained word embedding model. Unlike ZOD, we aim at mitigating pixel-level biased classification (semantic segmentation) rather than region-level biased classification (object detection).

The cornerstone of our solution to this problem is to develop both class-specific and class-agnostic prototypes in the training episodes and thus construct \textit{complete} feature pairs for comparison, as shown in Fig.~\ref{fig:compareFeature}(d). For the class-specific prototype, the model takes the foreground features in the query as positive samples and the background features as negative samples. In contrast, for each class-agnostic prototype, the corresponding background features in the query are treated as positive samples while the foreground features are viewed as negative samples. Note that the background is defined as any area outside those annotated objects (foreground). In this way, we can mitigate the prior bias that remembers novel-class objects as background in the training set. Specifically, the class-specific prototype is generated by using the support feature maps and mask annotations; and the class-agnostic prototypes are adaptively generated on the background features of query images.  Unlike previous methods~\cite{simpropnet, panet, pmms} that generate class-agnostic prototypes from the support images, we extract them from the query features alone to ensure the semantic similarity between these prototypes and the background features of query images. Finally, feature comparisons are conducted not only between the class-specific prototype and query features, but also between the class-agnostic prototypes and query features.

\begin{figure*}[tb]
\centering
\centerline{\includegraphics[width=.98\linewidth]{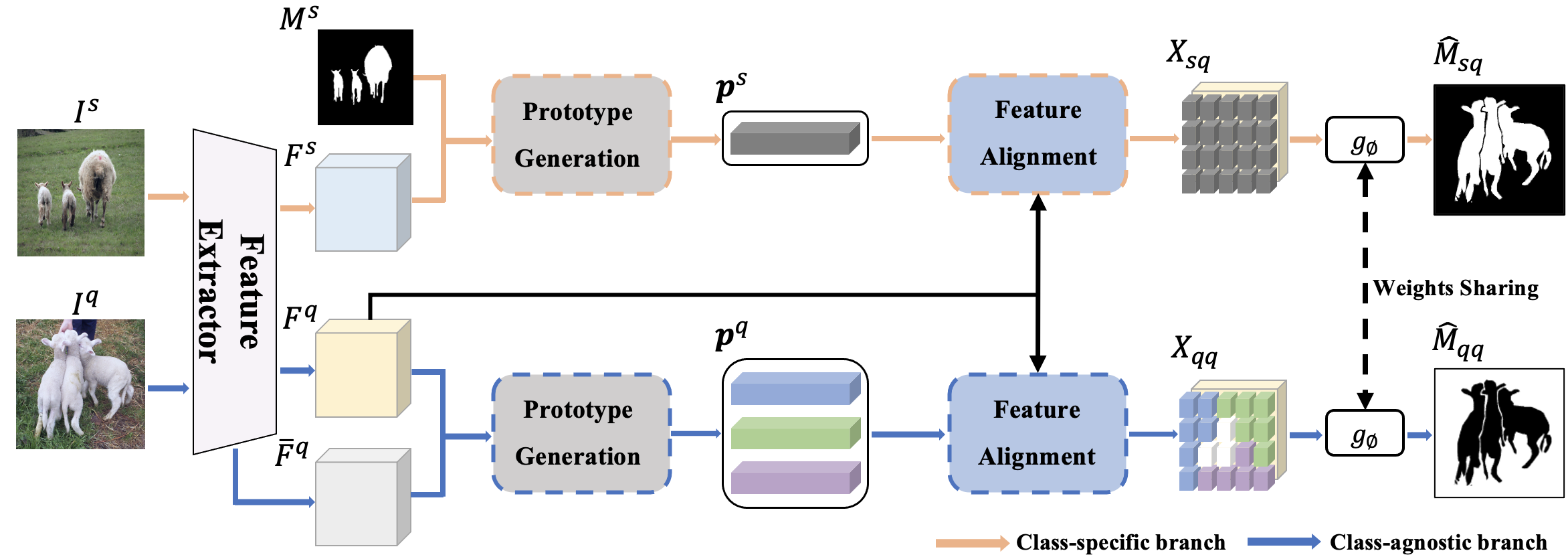}}
\caption{The pipeline of our APANet. APANet learns complementary feature comparison with two parallel branches, \ie, a class-specific branch and a class-agnostic branch. The class-specific and class-agnostic prototypes are firstly obtained by their individual prototype generation modules. Then, the feature comparison is conducted through feature alignment and a shared convolution module. Note that, between the two branches, the implementation is very different for prototype generation and feature alignment.}
\label{fig:overview}
\end{figure*}

The pipeline of our proposed APANet is shown in Fig.~\ref{fig:overview}. 
It consists of two parallel branches: a class-specific branch and a class-agnostic branch. Each branch has two sub-modules for prototype generation and feature alignment, respectively. For the class-specific branch, the class-specific prototype is obtained from the support images and aligned with the query features. Its goal is to segment the foreground area in the query image. In contrast, for the class-agnostic branch, we firstly use a simple clustering algorithm, $k$-means, to generate multiple class-agnostic prototypes with high-level query features. Then, it encourages each class-agnostic prototype to pull corresponding query features closer and push the foreground features away. This class-agnostic branch is \emph{not} involved in inference, hence it brings no extra computational cost to testing. In this way, the proposed APANet effectively and efficiently improves the FSS performance without additional parameters and computational cost.

We summarize the contributions of our work as follows.
\squishlist 
\item We present a novel learning paradigm for few-shot semantic segmentation. It learns feature comparison not only between the class-specific prototype and the query features (in the class-specific branch), but also between the class-agnostic prototypes and the query features (in the class-agnostic branch). To our best knowledge, we are the first to propose such a complementary features learning manner to help yield an unbiased segmentation model in the few-shot setting.

\item We propose a simple yet effective class-agnostic branch that includes class-agnostic prototypes generation and feature alignment. In practice, this branch can adaptively generate multiple class-agnostic prototypes from the background of a query image and learn feature alignment in a self-contrastive manner.

\item We achieve new state-of-the-art results on both the PASCAL-5$^i$ and COCO-20$^i$ datasets without additional parameters and computational cost at the inference stage. We also extensively demonstrate the effectiveness of the proposed method.
\squishend

\section{Related Work}
\noindent\textbf{Semantic Segmentation}  
Semantic segmentation aims to assign class labels to each image pixel, where labels are from a predefined set of semantic categories. Recent significant advances of this field have been made by using fully convolutional network~(FCN)~\cite{fcn}, which replaces the fully-connected layer in the network with convolutional layers. SegNet~\cite{segnet} introduces the encoder-decoder structure to be efficient in terms of both memory and computational time during inference. DeepLab~\cite{deeplab} adopts atrous spatial pyramid pooling (ASPP) to enlarge the receptive field of convolution layers, which is also widely used in few-shot segmentation~\cite{canet, pfenet, ppnet}. EMANet~\cite{emanet} proposes an Expectation-Maximization Attention module to reduce the computational complexity of self-attention and we also use such an online iterative method~($k$-means) at each step like them. However, these methods work for only a small number of fixed object classes and require a large number of image-mask pairs for training.

\noindent\textbf{Metric learning}  
Deep metric learning (DML) methods aim to learn an embedding space where instances from the same class are encouraged to be closer than those from different classes. Contrastive loss~\cite{hadsell2006dimensionality}, one of the classic metric learning methods, learns a discriminative metric via Siamese networks. It explicitly compares pairs of image representations to push away representations~(negative pair) from instances of different categories while pulling together those~(positive pair) from instances of the same categories. Triplet loss~\cite{tripletLoss} requires the similarity of a positive pair to be higher than that of a negative pair by a given margin. DML has been applied to various tasks including the few-shot segmentation. Most existing works, \eg,~\cite{PGNet, canet, pfenet, PL, pmms}, commonly extract the foreground prototype only, and then learn to pull the foreground features in the query closer to the prototype and push the background features away from it. This means that only the foreground features in the query are viewed 
as positive samples while the background features are treated as negative samples. \emph{In our work}, we construct class-specific and class-agnostic prototypes so that the foreground (background) features in the query can serve as positive or negative samples.

\noindent\textbf{Few-Shot Segmentation}  
The approach of few-shot learning is to train a network
on the training set and fine-tune it with the few data of
the novel classes. Such scenarios can be applied to many tasks, such as video classification~\cite{TMM_video_recog}, image recognition~\cite{TMM_imgage_recog1, TMM_imgage_recog2}, image generation~\cite{TMM_image_generation} or image semantic segmentation~\cite{HFA, OSLSM, pfenet}.

Few-shot semantic segmentation aims to perform pixel-level classification for novel classes in a query image conditioned on only a few annotated support images. OSLSM~\cite{OSLSM} first introduces this setting and uses parametric classification to solve this problem. PL~\cite{PL} and PANet~\cite{canet} use prototypes to represent typical information for the foreground objects present in the support images and make predictions by pixel-level feature comparison between the prototypes and query features via cosine similarity. This comparison can also be performed implicitly. For example, CANet~\cite{canet} uses convolution to replace the cosine similarity for complex objects variation in query images. Following this architecture, PFENet~\cite{pfenet} further exploits multi-scale query features to strengthen its representation ability. Recently, many studies try to fully mine foreground features from the support images. PGNet~\cite{PGNet} proposes a graph attention unit that treats each location of the foreground features in the support images as an individual and establishes the pixel-to-pixel correspondence between the query and support features. PMMs~\cite{pmms} uses the prototype mixture model to correlate diverse image regions with multiple prototypes. However, these works use only the foreground~(from the support images) to guide segmentation on the query images, as shown in Fig.~\ref{fig:compareFeature}(a). In addition, some works~\cite{simpropnet, panet} try to take into account the background information, but they only generate background prototypes with the support images, which is different from ours. \emph{In our work}, we explicitly generate multiple background prototypes from query images and learn feature alignment in a self-contrastive manner.

\noindent\textbf{Self-Supervised Learning}
In recent years, self-supervised learning has made remarkable success in unsupervised representation learning. It aims at designing pretext tasks to generate pseudo labels without additional manual annotations. Typical pretext tasks include predicting the angle of object rotation~\cite{gidaris2018unsupervised,feng2019self}, coloring a grayscale~\cite{zhang2016colorful}, inpainting missing regions~\cite{pathak2016context} and solving jigsaw puzzles~\cite{noroozi2016unsupervised}. An alternative line of works use clustering for unsupervised learning~\cite{SeLa2020,Cliquecnn2016,Deepclustering2018,caron2019unsupervised, xie2016unsupervised}, which use cluster assignments as pseudo-labels to learn deep representations. Deep clustering~\cite{Deepclustering2018} show that $k$-means assignments can be used as pseudo-labels to learn visual representations. This method scales to a large uncurated dataset and can be used for pre-training of supervised networks~\cite{caron2019unsupervised}. SeLa~\cite{SeLa2020} takes a pseudo-label assignment as an optimal transport problem and conducts simultaneous clustering and representation learning. Recently, contrastive learning has been widely used in self-supervised learning and currently achieved state-of-the-art performance on image classification~\cite{BYOL, MoCo, simclr}. The goal of contrastive representation learning is to learn an embedding space in which similar sample pairs stay close to each other while dissimilar ones are far apart. 

\emph{Our work} is related to clustering and contrastive learning. Similarly to deep clustering~\cite{Deepclustering2018}, we also use $k$-means clustering to generate class-agnostic prototypes (clustering centroids) with the high-level semantic features of query images. Then, dense feature comparisons are conducted between these class-agnostic prototypes and the query features by the class-agnostic branch, and the learning goal is to pull each class-agnostic prototype closer to the corresponding background features and push it away from the foreground features in the query.

\section{Our Method}
In this section, we first define the few-shot semantic segmentation task. Then, we describe the class-specific branch in Section~\ref{sec:SQP}. The details of the proposed class-agnostic branch are presented in Section~\ref{sec:QQP}. Finally, Section~\ref{sec:loss} describes the optimization and inference of our model.

\subsection{Preliminary}
Few-shot semantic segmentation (FSS) aims to segment the area of unseen class $C_{novel}$ from each query image given few labeled support images. Models are trained on base classes $C_{base}$ (\emph{training set}) and tested on novel classes $C_{novel}$ (\emph{test set}). Notice that $C_{base}$ and $C_{novel}$ are non-overlapping (\ie, $C_{base}$$\cap$$C_{novel}$=$\phi$), 
which ensures that the generalization ability of segmentation model to new class can be evaluated.

We adopt the episode training mechanism, which has been demonstrated as an effective approach to few-shot learning. Each episode is composed of a support set $S$ and a query set $Q$ of the same classes. The support set $S$ has $k$ image-mask pairs, \ie, $S=\{(I_{i}^{s},M_{i}^{s} )\}_{i=1}^{k}$, which is termed ``$k$-shot", with the $i$-th support image $I_{i}$ and its corresponding mask $M_i$. For query set $Q=\{(I^q,M^q)\}$, where $I^q$ is a query image and $M^q$ is its ground truth mask. The support-query triples $(\{(I_{i}^{s},M_{i}^{s} )\}_{i=1}^{k}, I^q)$ form the input data of FSS model, and the goal is to maximize the similarity between $M^q$ and the generated prediction $\hat{M}^q$ on $I^q$. Therefore, how to exploit $S$ for the segmentation of $I^q$ is a key task of few-shot segmentation.

To simplify the notation, let us take the ``1-shot" segmentation for example, \ie, $S=(I^{s},M^{s})$. That is, given a triple $(I^{s},M^{s}, I^q)$ as input, the goal of our proposed APANet is to directly learn a conditional probability mass function $\hat {M}^q = p(M^q|(I^{s},M^{s}, I^q); \vec \theta_f, \vec \theta_l)$ on base set $C_{base}$, where $\vec\theta_f$ represents the backbone parameters and $\vec\theta_l$ denotes the learnable parameters of the whole network. We use $(F^s, F^q)$ to denote the feature map of $(I^s, I^q)$ from a CNN backbone, \eg, VGG-Net or ResNet, where $F^s$ and $F^q \in \mathbb{R}^{h\times w \times c}$.

\subsection{Class-specific Branch: compare support and query}\label{sec:SQP}
The class-specific branch takes the foreground prototype of support images to compare it with the query features, whose objective is to correctly segment all query pixels, which belong to the foreground class as the foreground prototype or the background class as opposite the foreground prototype.
It consists of two sub-modules, prototype generation and feature alignment. In the prototype generation module, the class-specific prototype $\vec p^s$ 
is generated by mask average pooling (MAP) over the support features at locations $(i,j)$: 
\begin{equation}\label{eq:map}
 \vec p^s = \frac {\sum_{i,j}{\vec F_{i,j}^s \odot M_{i,j}^s}}{\sum_{i,j}{M_{i,j}^s}},
\end{equation}
where $\odot$ is the broadcast element-wised product. The prototype represents a specific semantic class present in the support image. Then, in the feature alignment module, the prototype $\vec p^s$ is assigned to each spatial location on query features $F^q$ and learns feature comparison to identify foreground objects presented in the query image. Technically, we first expand $\vec p^s$ to the same shape as the query features $F^q$ and concatenate them along the channel dimension as
\begin{equation}\label{eq:Xsq}
X_{sq} = \mathcal{C}\big(\mathcal E^s(\vec p^s), F^q\big),
\end{equation}
where $\mathcal{E}$ is the expansion operation, and $\mathcal{C}$ is the concatenation operation between two tensors along the channel dimension.
In order to obtain a better segmentation mask $\hat M^q$ on query image $I^q$, we follow previous methods~\cite{canet, dan}, which use a convolution module $g_\phi$ to encode the fusion feature $X_{sq}$, as
\begin{equation}\label{eq:Msq}
\hat M_{sq}^q = g_\phi(X_{sq}; \vec \theta_l).  
\end{equation}
Here, $\vec \theta_l$ denotes the learnable parameters in this module, and $g_\phi$ performs the pixel-level feature comparison in the way of binary classification, which verifies  whether the pixels in the feature map $F^q$ match the prototypes in $\mathcal{E}^s(\vec{p}^s)$ at the corresponding location. 

\subsection{Class-agnostic Branch: compare query and query} \label{sec:QQP}
In parallel to the class-specific branch, the class-agnostic branch takes multiple background prototypes of the query image instead of a single foreground prototype of the support image to conduct feature comparisons. It also consists of two sub-modules like the class-specific branch, but they are more elaborate. In the prototype generation module, the class-agnostic branch adaptively divides the background of the query image into several regions and extracts corresponding prototypes from them. In the feature alignment module, it learns feature alignment between these class-agnostic prototypes and query features. 

\subsubsection{Class-agnostic Prototypes Generation} 
Note that the class-specific branch extracts a single prototype with the MAP on the support features and compares it with the query features, because there are the same-class foreground objects in both the support and query images. However, it does not follow this assumption that there is similar semantics in the background of support-query pairs. Thus we think that it is more appropriate to estimate background prototypes directly from the query features. Here, we use an implicit assumption that a sample must be the positive sample of itself.

To derive the background prototype, a naive method is to directly apply the MAP operation~(Eq.~\ref{eq:map}) on the background region of the query features. But as the background is usually much more diverse than the particular foreground, simple MAP is actually ineffective. To alleviate this issue, we propose to adaptively partition the query features spatially according to their semantic distribution in the high-level feature space.

\begin{figure*}[t]
\centering
\centerline{\includegraphics[width=1\linewidth]{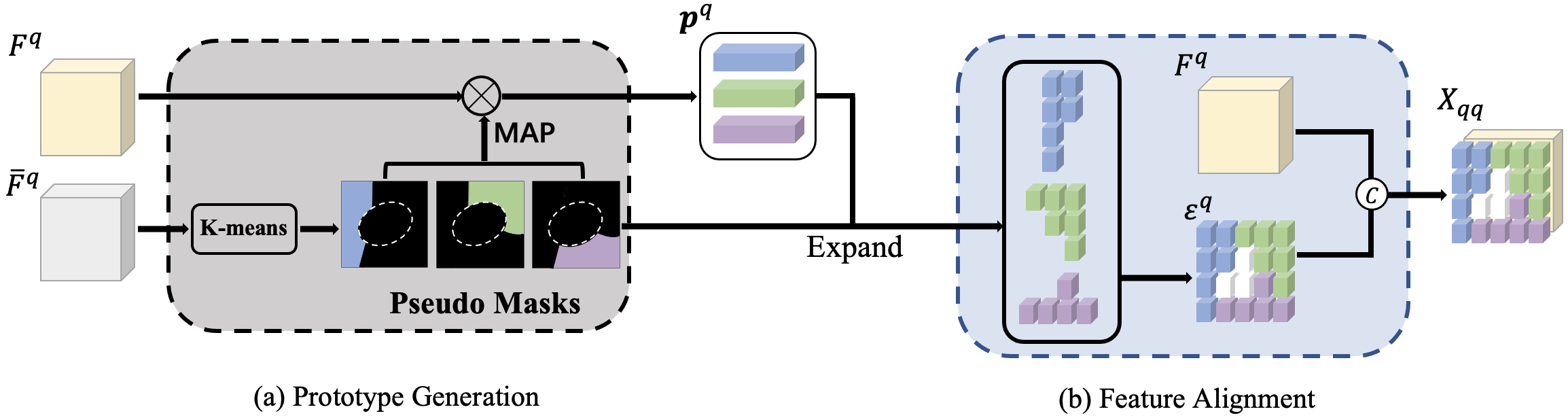}}
\caption{Illustration of our prototype generation and feature alignment modules in the class-agnostic branch. The query feature $F^q$ is first spatially partitioned into 3 regions by clustering on high-level feature $\bar {F}^q$. Then, the region prototypes $\vec p^q$ are generated by the mask average pooling (MAP). Finally, the prototypes are expanded to the corresponding region and concatenated with the query features to perform comparison. The white square in $\mathcal E^q$ represents randomly selected prototype from $\vec p^q$.}
\label{fig:module}
\end{figure*}

As we know, high-level semantic features are usually extracted by the deeper layers of a network. Therefore, instead of using the $F^q$ from the middle layers, we employ higher-level feature $\bar {F}^q$ to group the background of $F^q$. To partition $\bar {F}^q$ into $n$ regions, we use classical $k$-means clustering, the optimization problem of which can be expressed as 
\begin{align}\label{eq:kmeans}
\begin{split}
   & \min_{r_{ik}, \vec u_k} \sum_{i=1}^{hw} \sum_{k=1}^{n} r_{ik} {\rm dist}\left(\vec {\bar{F}}_{i}^{q}, \vec u_{k}\right),\\
&s.t.\ \ \sum_{k=1}^n{r_{ik}}=1, \forall i,
\end{split}
\end{align}
where ${\rm dist}(\cdot)$ is the standard cosine distance, the binary indicator variables $r_{ik}\in\{0,\ 1\}$, $\vec {\bar{F}}_{i}^{q} \in \bar F^q$, and $\vec u_k$ represents the centre of the $k$-th cluster. The optimal solution $r_{ik}^*$
and $\vec u_k^*$ can be readily obtained iteratively.

With $r_{ik}^*$, we can obtain the background prototype of $k$-th cluster region of $F^q$. First, we reshape the $k$-th binary indicator vector $\{r_{ik}\}_{i=1}^{hw}\in \mathbb{R}^{hw}$ to $\bar{M}_k^q\in \mathbb{R}^{h\times w}$. Hence, $\bar{M}_{k}^q$ depicts which spatial locations belong to the $k$-th cluster, and these locations are equal to one while others are zero. Notice that $\bar{M}_{k}^q$ may contain true foreground, so we check the intersection of $\bar{M}_k^q$ and the foreground mask ${M}^q$ and update $\bar{M_k}^q$ as    
\begin{equation}\label{eq:pm}
\bar{M_k}^q \gets \bar{M_k}^q-\bar{M_k}^q \cap {M}^q.
\end{equation}
The $\bar{M_k}^q$ represents the $k$-th mask that implies a potential semantic class presented in the background of query image, as shown in Fig~\ref{fig:KM_graph}. Then, we define a set of background prototypes to represent them. Considering that the specific categories of these background prototypes are not known, so we call them class-agnostic prototypes. Here, the $k$-th class-agnostic prototype is computed by using the MAP operation over the query features at locations $(i, j)$:
\begin{equation}\label{eq:bpmap}
\vec p_{k}^{q}= \frac {\sum_{i, j}{\vec {F_{i, j}^q} \odot  \bar M_{i,j,k}^q}}{\sum_{i, j}{ \bar M_{i,j,k}^q}},
\end{equation}
where $k=1, 2, \ldots, n$. 

\subsubsection{Feature Alignment for Complete Comparison} 
As discussed above, the class-specific branch pushes the background features in the query image away from the foreground prototype and encourages the foreground features in the query image to be close to the foreground prototype. Therefore, this learning mechanism tends to force the model to \emph{remember} the objects outside the base class set $C_{base}$ as background during training, which limits the generalization ability of the model. In our work, using both class-specific and class-agnostic branches, we construct complete feature comparisons to mitigate this issue.

Now we have obtained $n$ class-agnostic prototypes from a query image. How to align these prototypes to query features $F^q$ in the class-agnostic branch? Note that it is reasonable that the foreground and background regions are assumed to be known since we can utilize the annotation information of the query image at the training stage (although not for a query image at the testing stage). Therefore, here the prototype assignment problem of query features can be decomposed into two aspects, for the foreground and background features in the query image, respectively. 

For the background features in the query image, considering that these class-agnostic prototypes are derived from the background region through clustering, each class-agnostic prototype (\eg, the $k$-th prototype $\vec p_k^q$) corresponds to a specific sub-region~($\bar{M}_{k}^q$). Therefore, it is natural that a prototype should be as close as possible to its corresponding background features and meanwhile as far as possible from the foreground features. Technically, we first expand each prototype $\vec p_k^q$ according to $\bar{M}_{k}^q$. Then, all the expanded prototypes $\vec p_k^q\  (k=1,2,\ldots, n)$ and their corresponding query features are concatenated along the channel dimension as in Eq.~(\ref{eq:Xqq}), as shown in Fig.~\ref{fig:module}(b). In this way, we construct positive pairs between these class-agnostic prototypes and all background features in the query image. 

Next, how to assign one prototype to the foreground features of the query image to construct negative pairs for the class-agnostic branch? If, similar to the class-specific branch, we directly assign the class-specific prototype to all foreground features in the query image, positive pairs are constructed between the class-specific prototype and foreground features in the query image since they belong to the same class. However, combined with the assignment strategy of background features of the query image, it will lead to a trivial solution because only positive pairs are constructed between all query features and these assigned prototypes. Therefore, we need to construct negative pairs through assigning suitable prototypes for the foreground features of the query image. To this end, we randomly select one class-agnostic prototype and densely pair it with each location of foreground features of the query image such that negative pairs are constructed. 

We denote the above expansion and assignment operation for all class-agnostic prototypes as $\mathcal E^q(\cdot)$, and similarly to Eq.~(\ref{eq:Xsq}) we have
\begin{equation}\label{eq:Xqq}
X_{qq}= \mathcal{C}\big(\mathcal E^q(\vec p_k^q|k=1,2,\cdot,n), F^q\big).
\end{equation}
Finally, a convolution module $g_\phi$ is used to encode $X_{qq}$ for learning the comparison metric as
\begin{equation}\label{eq:Mqq}
\hat M_{qq}^q = g_\phi(X_{qq}; \vec \theta_l).  
\end{equation}
Note that all parameters~(\ie, $\vec \theta_l$ in Eq.~\ref{eq:Msq} and Eq.~\ref{eq:Mqq}) are shared to learn an unbiased segmentation model. We develop class-specific and class-agnostic prototypes and thus construct complete feature pairs~(\ie, $X_{sq}$ and $X_{qq}$) to mitigate the biased classification issue for few-shot semantic segmentation. Only under the condition of parameter sharing, it is possible to learn an unbiased segmentation model, because the model can be trained with complete sample pairs including positive and negative ones.

\begin{algorithm}[t]
\caption{The training pipeline of APANet.}
\label{algorithm1}
\KwIn{Support feature $F^s$, support mask $M^s$, query feature $F^q$, high-level query feature $\bar {F}^q$, cluster number $K$}
\KwOut{Learnable parameters $\vec \theta_l$}
\tcp{compare between support and query}
generate class-specific prototype $\vec {p}^s$ with Eq.~\ref{eq:map};\\
expand and concatenate prototype with query: $X_{sq} = C(\mathcal E^s(\vec{p}^s), F^q)$ (Eq.~\ref{eq:Xsq});\\
compare with query: $\hat M_{sq}^q = g_\phi(X_{sq}; \vec \theta_l)$\ (Eq.~\ref{eq:Msq});\\
\tcp{compare between query and query}
cluster $\bar {F}^q$ with Eq.~\ref{eq:kmeans} and get $ \{\bar M_k^q \}_{k=1}^K$\;
\For{$k = 1, ..., K$}{
    $\bar{M_k}^q \leftarrow \bar{M_k}^q-\bar{M_k}^q \cap {M}^q $\ (Eq.~\ref{eq:pm});\\
    generate class-agnostic prototype $\vec {p_k}^q$ with Eq.~\ref{eq:bpmap}\;
    \For {each position $(i,j)$, where $M_{i,j,k}^q=1$}
    {
        expand $\vec {p_k}^q$ to the position $(i,j)$ of $F^q$, \\
        concatenate with query: \\
        $X_{qq}^k = \mathcal{C}(\mathcal{E}^q(\vec {p_k}^q), F^q)$\ (Eq.~\ref{eq:Xqq});
    }
}
compare with query: $\hat M_{qq} = g_\phi(X_{qq}; \vec \theta_l)$\ (Eq.~\ref{eq:Mqq});\\
\tcp{Compute loss}
compute loss $\mathcal L$ with Eq.~\ref{eq:loss}; \\
\tcp{Update parameters}
update parameters $\vec \theta_l$.
\end{algorithm}

\begin{algorithm}[t]
\caption{The inference pipeline of APANet.}
\label{algorithm2}
\KwIn{Support feature $F^s$, support mask $M^s$, query feature $F^q$ and learned model $g_\phi(\cdot;\vec \theta_l)$}
\KwOut{Prediction mask $\hat M_{sq}^q$ of query image $I^q$}
generate class-specific prototype $\vec {p}^s$ with Eq.~\ref{eq:map};\\
expand and concatenate prototype with query: $X_{sq} = C(\mathcal E^s(\vec{p}^s), F^q)$ (Eq.~\ref{eq:Xsq});\\
predict mask: $\hat M_{sq}^q = g_\phi(X_{sq}; \vec \theta_l)$\ (Eq.~\ref{eq:Msq}).\\
\end{algorithm}

\subsection{Optimization and Inference}\label{sec:loss}
\subsubsection{Loss Function}
It requires two different supervision signals to guide the learning of the class-specific and class-agnostic branches during training. Essentially, the learning of these two branches shares the same spirit to check each query pixel whether or not there is the same semantic class between the query feature and its aligned prototype, regardless of the class-specific or class-agnostic prototype. Hence, if the comparison is based on the same semantic (\ie, the class-specific prototype and the foreground features, or the class-agnostic prototype and the corresponding background features), the model should output consistent prediction~(\ie, 1) at these positions and 0 otherwise.

Note that the class-specific branch learns feature comparisons between the class-specific prototype and each query feature and finally outputs $\hat M_{sq}^q$. This branch should encourage activating the foreground region while suppressing the background area in the query image. Different from the class-specific branch, the class-agnostic branch learns feature comparisons between the class-agnostic prototypes and each query feature and finally outputs $\hat M_{qq}^q$. Therefore, the learning goal of the class-agnostic branch is to activate the background region while suppressing the foreground area of the query image. That is, for the class-specific branch, its target label is the mask $M^q$ of query image, while for the class-agnostic branch, its target label is $1-M^q$.
We use the cross-entropy loss and formulate the overall loss function as
\begin{equation}\label{eq:loss}
    \mathcal L=(1-\lambda) \mathcal L_{1}\left(\hat{M}_{sq}, M^{q}\right)+\lambda \mathcal L_{2}\left(\hat {M}_{qq}, 1-M^{q}\right),
\end{equation}
where $\lambda$ is a parameter to balance the two cross-entropy losses.
When $\lambda$ is 0, only the $\mathcal L_1$ is left in the overall loss function, then our APANet degenerates to the baseline. The whole training process of our APANet is delineated in Algorithm~\ref{algorithm1}.

\subsubsection{Inference}
Note that each one of our two branches can yield a segmentation result in the training phase. Unlike the training stage, the model just needs output $\hat {M}_{sq}$ by using the class-specific branch during the inference stage because the mask of a query image is now unknown. Hence, the class-agnostic branch can be freely removed, which brings no extra inference costs. Given a query image and $k$ support images, we take the average of all foreground prototypes from $k$ support images as the new foreground prototype. The prototype is then used to output $\hat {M}_{sq}$ as prediction. The whole inference process of our APANet is described in Algorithm~\ref{algorithm2}. Compared with the model without the class-agnostic branch~(inference stage), our model with both the class-specific and class-agnostic branches takes about 1.3$\times$ more time for training. The increased training time is mainly brought by our online clustering in the class-agnostic branch.

\begin{table*}[h]
\caption{Class mIoU results of 1-shot and 5-shot segmentation on PASCAL-5$^i$. Best results are in bold.}
\label{table:pascal}
\resizebox{\textwidth}{!}{
\begin{tabular}{@{}l|c|cccc|c|cccc|c@{}}
\toprule
\multirow{2}{*}{Methods}                     &\multirow{2}{*}{Backbone}         & \multicolumn{5}{c|}{1-Shot}                                                   & \multicolumn{5}{c}{5-Shot}                                                    \\ \cmidrule(l){3-12} 
              &  & Fold-0       & Fold-1       & Fold-2       & Fold-3       & Mean          & Fold-0       & Fold-1       & Fold-2       & Fold-3       & Mean          \\ \midrule
SG-One~\cite{sgone}(TCYB'20)      &          & 40.2          & 58.4          & 48.4          & 38.4          & 46.3          & 41.9          & 58.6          & 48.6          & 39.4          & 47.1          \\
AMP~\cite{amp}(ICCV'19)         &    & 41.9          & 50.2          & 46.7          & 34.7          & 43.4          & 41.8          & 55.5          & 50.3          & 39.9          & 46.9          \\
PANet~\cite{panet}(ICCV'19)       &          & 42.3          & 58.0          & 51.1          & 41.2          & 48.1          & 51.8          & 64.6          & 59.8          & 46.5          & 55.7          \\
RPMM~\cite{pmms}(ECCV'20)        & VGG-16          & 47.1          & 65.8          & 50.6          & 48.5          & 53.0          & 55.0          & 66.5          & 51.9          & 47.6          & 54.0          \\
FWB~\cite{fwb}(ICCV'19)         &          & 47.0          & 59.6          & 52.6          & 48.3          & 51.9          & 50.9          & 62.9          & 56.5          & 50.1          & 55.1          \\ 
PFENet~\cite{pfenet}(TPAMI'20)         &          & 56.9          & 68.2         & 54.4          & \textbf{52.4}          & 58.0          & 59.0          & 69.1           & 54.8         & 52.9          & 59.0          \\ 
APANet(ours)         &          & \textbf{58.0}          & \textbf{68.9}         & \textbf{57.0}          & 52.2          & \textbf{59.0}          & \textbf{59.8}          & \textbf{70.0}           & \textbf{62.7}         & \textbf{57.7}          & \textbf{62.6}          \\ \midrule
CANet~\cite{canet}(CVPR'19)       &          & 52.5          & 65.9          & 51.3          & 51.9          & 55.4          & 55.5          & 67.8          & 51.9          & 53.2          & 57.1          \\
PGNet~\cite{PGNet}(ICCV'19)       &          & 56.0          & 66.9          & 50.6          & 56.0          & 57.7          & 57.7          & 68.7          & 52.9          & 54.6          & 58.5          \\
CRNet~\cite{crnet}(CVPR'20)       &          & -             & -             & -             & -             & 55.7          & -             & -             & -             & -             & 58.8          \\
SimPropNet~\cite{simpropnet}(IJCAI'20) &  & 54.9          & 67.3          & 54.5          & 52.0            & 57.2          & 57.2          & 68.5          & 58.4          & 56.1          & 60.0           \\
RPMM~\cite{pmms}(ECCV'20)        & ResNet-50         & 55.2          & 66.9          & 52.6          & 50.7          & 56.3          & 56.3          & 67.3          & 54.5          & 51            & 57.3          \\
PFENet~\cite{pfenet}(TPAMI'20)     &          & 61.7          & 69.5          & 55.4          & 56.3          & 60.8          & 63.1          & 70.7          & 55.8          & 57.9          & 61.9          \\
SAGNN~\cite{SAGNN}(CVPR'21)     &          & \textbf{64.7}          & 69.6          & 57.0          & 57.2          & 62.1          & \textbf{64.9}          & 70.0          & 57.9          & 59.3          & 62.8          \\
SCLNet~\cite{SCLNet}(CVPR'21)     &          & 63.0          & 70.0          & 56.5          & 57.7          & 61.8          & 64.5          & 70.9          & 57.3          & 58.7          & 62.9          \\
APANet(ours)          &          & 62.2 & \textbf{70.5} & \textbf{61.1} & \textbf{58.1} & \textbf{63.0} & 63.3 & \textbf{72.0} & \textbf{68.4} & \textbf{60.2} & \textbf{66.0} \\ \midrule
FWB~\cite{fwb}(ICCV'19)     &\multirow{4}{*}{ResNet-101}           & 51.3          & 64.5          & 56.7          & 52.2          & 56.2          & 54.8          & 67.4          & 62.2          & 55.3          & 59.9         \\ 
DAN~\cite{dan}(ECCV'20)     &          & 54.7          & 68.6          & 57.8          & 51.6          & 58.2          & 57.9          & 69.0          & 60.1          & 54.9          & 60.5         \\ 
PFENet~\cite{pfenet}(TPAMI'20)     &         & 60.5          & 69.4          & 54.4          & 55.9          & 60.1          & 62.8          & 70.4          & 54.9          & 57.6          & 61.4          \\
APANet(ours)     &          & \textbf{63.1}          & \textbf{71.1}          & \textbf{63.8}          & \textbf{57.9}          & \textbf{64.0}         & \textbf{67.5}          & \textbf{73.3}          & \textbf{67.9}          & \textbf{63.1}          & \textbf{68.0}          \\
\bottomrule
\end{tabular}}
\end{table*}

\begin{table*}[t]
\caption{Class mIoU/FB-IoU results of 1-shot and 5-shot segmentation on COCO-20$^i$. Best results are in bold.}
\label{coco}
\resizebox{\textwidth}{!}{
\begin{tabular}{@{}l|c|cccc|c|cccc|c@{}}
\toprule
\multirow{2}{*}{Methods}                 &\multirow{2}{*}{Backbone}          & \multicolumn{5}{c|}{1-Shot}                                                 & \multicolumn{5}{c}{5-Shot}                            \\ \cmidrule(l){3-12} 
           &   & Fold-0 & Fold-1 & Fold-2 & Fold-3 & \multicolumn{1}{c|}{Mean}          & Fold-0 & Fold-1 & Fold-2 & Fold-3 & Mean          \\ \midrule
          \multicolumn{12}{c}{Class mIoU Evaluation}\\
\hline
FWB~\cite{fwb}(ICCV'19)     &     & 18.4    & 16.7    & 19.6    & 25.4    & \multicolumn{1}{c|}{20.0}          & 20.9    & 19.2    & 21.9    & 28.4    & 22.6          \\ 
PANet~\cite{panet}(ICCV'19)     &     & -    & -    & -    & -    & \multicolumn{1}{c|}{20.9}          & -    & -    & -    & -    & 29.7          \\
PFENet~\cite{pfenet}(TPAMI'20)     & VGG-16    & 33.4    & 36.0    & 34.1    & 32.8    & \multicolumn{1}{c|}{34.1}          & 35.9    & 40.7    & 38.1    & 36.1    & 37.7          \\
SAGNN~\cite{SAGNN}(CVPR'21)     &     & 35.0    & \textbf{40.5}    & \textbf{37.6}    & 36.0    & \multicolumn{1}{c|}{\textbf{37.3}}          & 37.2   & 45.2    & 40.4    & 40.0    & 40.7          \\
APANet(ours)    &           & \textbf{35.6}    & {40.0}    & {36.0}    & \textbf{37.1}    & \multicolumn{1}{c|}{{37.2}}      & \textbf{40.1}    & \textbf{48.7}    & \textbf{43.3}    & \textbf{40.7}    & \textbf{43.2}          \\
\midrule
PPNet~\cite{ppnet}(ECCV'20)   &   & 34.5    & 25.4    & 24.3    & 18.6    & \multicolumn{1}{c|}{25.7}          & \textbf{48.3}    & 30.9    & 35.7    & 30.2    & 36.2          \\
RPMM~\cite{pmms}(ECCV'20)    & ResNet-50          & 29.5    & 36.8    & 29.0    & 27.0    & \multicolumn{1}{c|}{30.6}          & 33.8    & 42.0    & 33.0    & 33.3    & 35.5          \\ 
APANet(ours)    &           & \textbf{37.5}    & \textbf{43.9}    & \textbf{39.7}    & \textbf{40.7}    & \multicolumn{1}{c|}{\textbf{40.5}}      & 39.8    & \textbf{46.9}    & \textbf{43.1}    & \textbf{42.2}    & \textbf{43.0}          \\ 
\midrule
FWB~\cite{fwb}(ICCV'19) &\multirow{5}{*}{ResNet-101}  & 19.9    & 18.0    & 21.0    & 28.9    & \multicolumn{1}{c|}{21.2}          & 19.1    & 21.5    & 23.9    & 30.1    & 23.7          \\
PFENet~\cite{pfenet}(TPAMI'20) &  & 34.3    & 33.0    & 32.3    & 30.1    & \multicolumn{1}{c|}{32.4}          & 38.5    & 38.6    & 38.2    & 34.3    & 37.4          \\
SAGNN~\cite{SAGNN}(CVPR'21)     &     & 36.1    & 41.0    & 38.2    & 33.5    & \multicolumn{1}{c|}{37.2}          & 40.9   & 48.3    & 42.6    & 38.9    & 42.7          \\
SCLNet~\cite{SCLNet}(CVPR'21)     &     & 36.4    & 38.6    & 37.5    & 35.4    & \multicolumn{1}{c|}{37.0}          & 38.9   & 40.5    & 41.5    & 38.7    & 39.9          \\
APANet(ours)      &           & \textbf{40.7}    & \textbf{44.6}    & \textbf{42.5}    & \textbf{39.6}    & \multicolumn{1}{c|}{\textbf{41.9}} & \textbf{45.7}    & \textbf{49.7}    & \textbf{47.4}    & \textbf{42.8}    & \textbf{46.4} \\ 

\hline
\multicolumn{12}{c} {FB-IoU Evaluation}\\
\hline
{PANet~\cite{panet}(ICCV'19)}  & \multirow{4}{*}{VGG-16}          & -      & -      & -      & -      & 59.2 & -      & -      & -      & -      & 63.5 \\
{PFENet~\cite{pfenet}(TPAMI'20)} &     & 50.0   & 63.1   & 63.5   & 63.4   & 60.0 & 50.3   & 65.2   & 65.2   & 65.5   & 61.6 \\
SAGNN~\cite{SAGNN}(CVPR'21)     &      & -    & -    & -    & -    & \multicolumn{1}{c|}{61.2}          & -   & -    & -    & -    & 63.1          \\
APANet(ours)   &           & \textbf{50.5}        & \textbf{64.6}        & \textbf{64.7}        & \textbf{66.4}       & \textbf{61.6}   & \textbf{54.1}   & \textbf{70.5}   & \textbf{69.2}  & \textbf{70.0} & \textbf{66.0} \\
 \midrule
{A-MCG~\cite{a-mcg}(AAAI'19)}  &\multirow{2}{*}{ResNet-50}    & -      & -      & -      & -      & 59.2 & -      & -      & -      & -      & 63.5 \\
APANet(Ours)   &         & 50.1   & 67.9   & 67.8   & 68.6   & \textbf{63.6} & 56.7   & 69.7   & 70.2   & 70.1   & \textbf{66.7} \\ \midrule
{PFENet~\cite{pfenet}(TPAMI'20)} & & 52.2   & 59.5   & 61.5   & 61.4   & 58.6 & 51.5   & 65.6   & 65.7   & 64.7   & 61.9 \\
SAGNN~\cite{SAGNN}(CVPR'21)     &{ResNet-101}      & -    & -    & -    & -    & \multicolumn{1}{c|}{60.9}          & -   & -    & -    & -    & 63.4          \\
APANet(ours)   &           & \textbf{54.0}        & \textbf{67.9}        & \textbf{68.9}        & \textbf{68.3}       & \textbf{64.8}   & \textbf{63.6}   & \textbf{71.4}   & \textbf{72.2}  & \textbf{72.0} & \textbf{69.8} \\
\bottomrule
\end{tabular}}
\end{table*}

\begin{table}
\centering
\caption{FB-IoU results on PASCAL-5$^i$. As many other
methods do not report the specific result of each split, we present the comparison of the average FB-IoU results in this table.}
\label{FB_PASCAL}
\begin{tabular}{@{}ccc@{}}
\toprule
\multicolumn{1}{c|}{Methods} & \multicolumn{1}{c|}{1-Shot}        & 5-Shot        \\ \midrule
\multicolumn{3}{c}{VGG-16}                                                        \\ \midrule
\multicolumn{1}{c|}{co-FCN~\cite{co-fcn}}  & \multicolumn{1}{c|}{60.1}          & 60.2          \\
\multicolumn{1}{c|}{PL~\cite{PL}(BMVC'18)}      & \multicolumn{1}{c|}{61.2}          & 62.3          \\
\multicolumn{1}{c|}{SG-One~\cite{sgone}(TCYB'20)}  & \multicolumn{1}{c|}{63.9}          & 65.9          \\
\multicolumn{1}{c|}{PANet~\cite{panet}(ICCV'19)}   & \multicolumn{1}{c|}{66.5}          & 70.7          \\
\multicolumn{1}{c|}{PFENet~\cite{pfenet}(TPAMI'20)}  & \multicolumn{1}{c|}{\textbf{72.0}} & 72.3          \\
\multicolumn{1}{c|}{Ours}    & \multicolumn{1}{c|}{71.3}          & \textbf{75.2} \\ \midrule
\multicolumn{3}{c}{ResNet-50}                                                     \\ \midrule
\multicolumn{1}{c|}{CANet~\cite{canet}(CVPR'19)}   & \multicolumn{1}{c|}{66.2}          & 69.6          \\
\multicolumn{1}{c|}{PFENet~\cite{pfenet}(TPAMI'20)}  & \multicolumn{1}{c|}{\textbf{73.3}} & 73.9          \\
\multicolumn{1}{c|}{SAGNN~\cite{SAGNN}(CVPR'21)}  & \multicolumn{1}{c|}{73.2} & 73.3          \\
\multicolumn{1}{c|}{SCLNet~\cite{SCLNet}(CVPR'21)}  & \multicolumn{1}{c|}{71.9} & 72.8          \\
\multicolumn{1}{c|}{Ours}    & \multicolumn{1}{c|}{72.3}          & \textbf{77.4} \\ \midrule
\multicolumn{3}{c}{ResNet-101}                                                    \\ \midrule
\multicolumn{1}{c|}{A-MCG~\cite{a-mcg}(AAAI'19)}   & \multicolumn{1}{c|}{61.2}          & 62.2          \\
\multicolumn{1}{c|}{PFENet~\cite{pfenet}(TPAMI'20)}  & \multicolumn{1}{c|}{72.9}          & 73.5          \\
\multicolumn{1}{c|}{Ours}    & \multicolumn{1}{c|}{\textbf{74.9}} & \textbf{78.8} \\ \bottomrule
\end{tabular}
\end{table}

\section{Experiments}  
\subsection{Experiment Setting}
\noindent\textbf{Datasets}
We evaluate the proposed approach on two public few-shot segmentation benchmarks: PASCAL-5$^i$~\cite{OSLSM} and COCO-20$^i$~\cite{fwb, panet}.
PASCAL-5$^i$ is built from PASCAL VOC 2012~\cite{voc} and with extended annotations from SDS~\cite{sds}. This dataset contains 20 object classes divided into four folds and each fold has 5 categories. Following PFENet~\cite{pfenet}, 5,000 support-query pairs were randomly sampled in each test fold for evaluation. We also evaluate our approach on a more challenging dataset COCO-20$^i$, which is built on MS-COCO. It contains more samples, more classes and more instances per image. Following~\cite{fwb}, COCO-20$^i$ is also split into four folds from 80 classes and each fold contains 20 classes. We use the same class division and randomly sample 20,000 support-query pairs for evaluation, as done with PFENet~\cite{pfenet}.

For both datasets, we adopt 4-fold cross-validation that trains model on three folds~(base class) and tests on the remaining fold (novel class). The experimental results are reported on each test fold. We also report the average performance of all four test folds.

\noindent\textbf{Evaluation Metric}
Following~\cite{pfenet, fwb, canet}, we use the widely adopted mean intersection over union~({mIoU}) and the foreground-background IoU~(FB-IoU) for quantitative evaluation. For each class, the {IoU} is calculated by $\frac{\mathrm{TP}}{\mathrm{TP}+\mathrm{FP}+\mathrm{FN}}$, where {TP} is the number of true positives, {FP} is the number of false positives and {FN} is the number of false positives over the prediction and ground-truth masks on the query set. The {mIoU} is an average of {IoU}'s over all classes, \ie, $\mathrm{mIoU}$=$\frac{1}{n_{c}} \sum_{i} \mathrm{IoU}_{i}$, where ${n_{c}}$ is the number of novel classes. FB-IoU treats all object classes as one foreground class and averages the IoU's of foreground and background~($C=2$). We take average of results on all folds as the final mIoU/FB-IoU. It is worth noting that FB-IoU is biased towards and benefits from the background class because most foreground classes only occupy a small spatial region of the whole image. Hence, the mIoU is the key evaluation criterion for FSS. In order to compare with the previous methods, we also report the FB-IoU results.

\subsection{Implementation Details}
We use a modified version of VGG-16~\cite{vgg}, ResNet-50 or ResNet-101~\cite{resnet} as the backbone for fair comparison with other methods. These backbone networks are initialized with ImageNet pre-trained weights and we keep their weights~($\vec \theta_f$) fixed during training. Other layers~($\vec \theta_l$) are initialized by the default setting of PyTorch. The ResNet we use is the dilated version used in previous work~\cite{pfenet} and the VGG we use is the original version~\cite{vgg}. In ResNet and VGG, there are 5 blocks which correspond to 5 different levels of feature representations. We extract pre-trained features from the third and fourth blocks and concatenate them to generate image features $F^s$ and $F^q$. The high-level prior feature $\bar F^q$ is obtained from the last block. Similarly to PFENet, we also use a multi-scale Feature Enrichment Module to incorporate the prior masks to enrich the query features in Eq.~\ref{eq:Xsq} and Eq.~\ref{eq:Xqq}. The comparison function $g_\phi$ shares the same parameter in different branches, consisting of a 3$\times$3 convolution and a 1$\times$1 convolution with 2 output channels~(background and foreground). 

The network is trained on PASCAL-5$^i$ with the initial learning of 0.0025 and the momentum of 0.9 for 200 epochs with 4 pairs of support-query images per batch. For COCO-20$^i$, models are trained for 50 epochs with a learning rate of 0.005 and batch size 8. We randomly crop 473$\times$473 patches from the processed images as the training samples. The $k$-means algorithm iterates for 10 rounds to calculate the pseudo mask for the query image. Data augmentation strategies including normalization, mirror operation and random rotation from -10 to 10 degrees are used. Our experiments do not use any post-processing techniques, such as CRF~\cite{crf}, to refine the results. All experiments are conducted on NVIDIA Tesla V100 GPUs and Intel Xeon CPU Platinum 8255C.

\begin{figure*}[t]
\centering
\centerline{\includegraphics[width=1\linewidth]{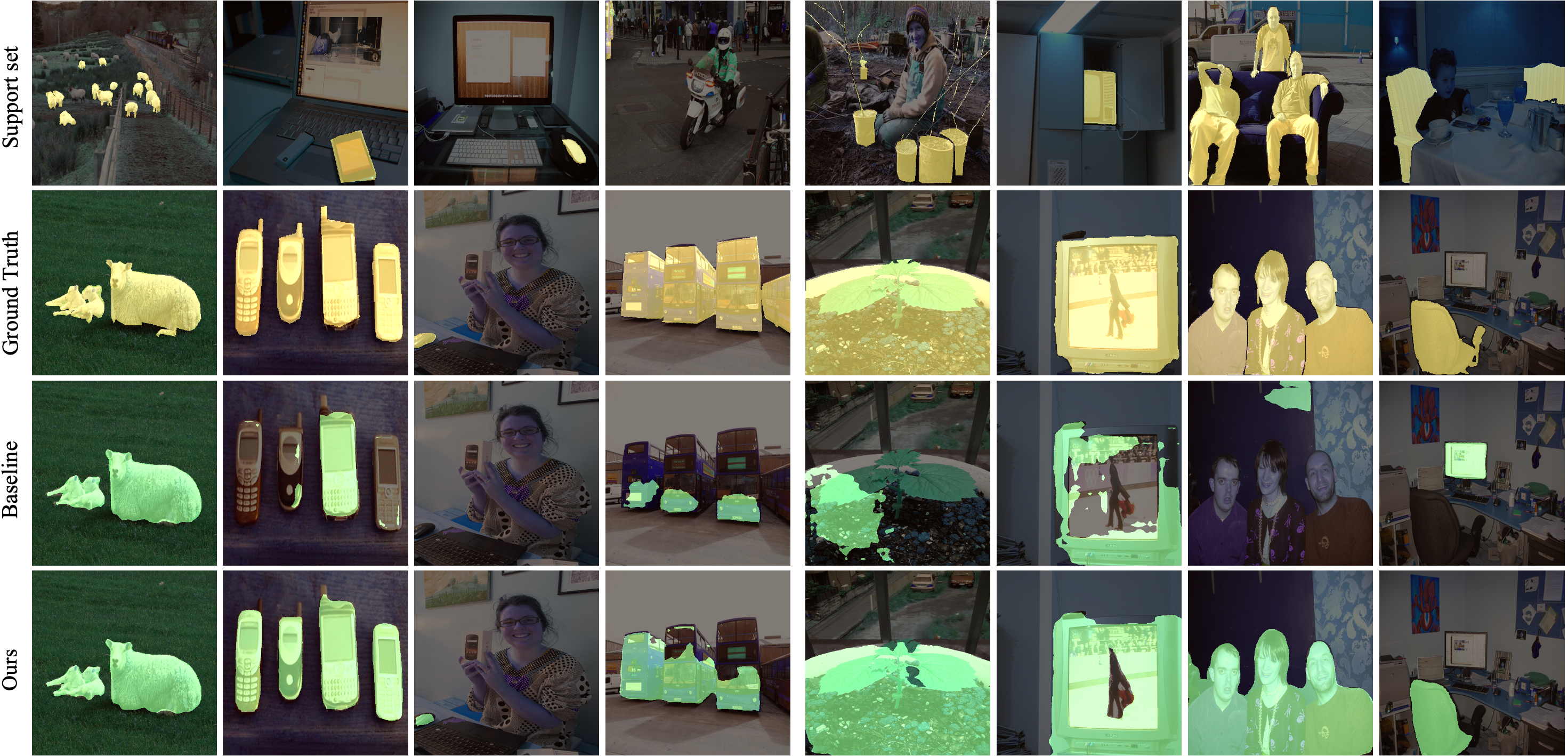}}
\caption{Qualitative results of the proposed APANet and the baseline. The left samples are from COCO-20$^i$ and the right ones are from PASCAL-5$^i$. The first and second rows are for the support and query images with their ground-truth annotations. The third and fourth rows are for the segmentation results of the baseline and our APANet.}
\label{fig:qvis}
\end{figure*}

\subsection{Comparison with State-of-the-Arts} 
As reported in Tables~\ref{table:pascal},~\ref{coco} and~\ref{FB_PASCAL},
we compare the proposed method with the state-of-the-arts on PASCAL-5$^i$ and COCO-20$^i$ by using mIoU and FB-IoU. Our method significantly outperforms the state-of-the-arts in both the 1-shot and 5-shot settings. Additional qualitative results are shown in Fig.~\ref{fig:qvis}.

\noindent\textbf{PASCAL-5$^i$ Results.} 
We report the {mIoU} of each fold and the mean of all four folds on PASCAL-5$^i$ in Table~\ref{table:pascal}. We can see that our APANet significantly outperforms the state-of-the-arts with all backbones. Specifically, in the 1-shot setting, our method surpasses the state-of-the-arts by 0.9\% and 3.9\% with the ResNet-50 and ResNet-101 backbones, respectively. This indicates that exploiting complete feature comparison is clearly beneficial for few-shot semantic segmentation. In the 5-shot setting, our method significantly improves the state-of-the-arts by 3.1\% (62.9\% vs. 66.0\%) and 6.6\% (61.4 vs. 68.0\%). This further demonstrates its effectiveness for multi-shot cases. In Table~\ref{FB_PASCAL}, we further make comparisons in terms of FB-IoU. Equipped with the ResNet-101 backbone, our APANet obtains competitive top-performing 1-shot and 5-shot results. Our result is slightly weak in the 1-shot setting using the VGG-16 and ResNet-50 backbones. Nevertheless, we note that the FB-IoU criterion is biased towards and benefits from the background class.

\noindent\textbf{COCO-20$^i$ Results.}
The results on COCO-20$^i$ are reported in Table~\ref{coco}. Our method performs competitively with state-of-the-art approaches in the 1-shot setting and significantly outperforms recent methods in the 5-shot scenario. We still achieve the state-of-the-art results, with significant improvement (4.9\% and 6.3\% with 1- and 5-shot) even compared with the latest SCLNet~\cite{SCLNet} with the ResNet-101 backbone. Using the FB-IoU metric, the proposed APANet also achieves significant improvement over the state-of-the-arts in both the 1-shot and 5-shot settings. These results suggest that our additional class-agnostic branch promotes the model to learn better feature representation and yields an unbiased classifier, while other methods may tend to mis-predict some novel-class objects as the background even if more support information is supplied.

\noindent\textbf{Qualitative Results.} 
We show some qualitative results on the PASCAL-5$^i$ and COCO-20$^i$ test sets in Fig.~\ref{fig:qvis}. We can make the following observations. First, our method is capable of making correct predictions even if the background of the query image contains some other targets, \eg, the \emph{person} and \emph{TV} in the third and eighth columns. Second, the baseline method may segment only parts of the novel-class object because of the prior bias, which hinders the segmentation of the novel-class object. In contrast, the proposed method has a better generalization performance when there is diverse semantics among objects in the support-query images. Third, we note that our method may fail when the support objects are mostly occluded~(\eg, the bus in the fourth column) or some small objects are presented in the foreground~(\eg, the person on TV in the sixth column).

\subsection{Ablation Studies}  
In order to comprehend how APANet works, we perform extensive ablation studies to analyze the components in APANet. All experiment results are evaluated over all folds of the PASCAL-5$^i$ by using the ResNet-50 backbone.

\noindent\textbf{Number of Class-agnostic Prototype.} 
We fix the loss weight $\lambda$ to 0.5 and choose the value
$n$ (the number of clusters) from a given set $\{1, 2, 3, 4, 5\}$. As shown in Table~\ref{Number of $n$}, first, our method performs better than the baseline under all cases even in the simplest case~(\ie, $n=1$). Second, the best performance is achieved when $n$ is set to 3, which validates the introduction of clustering to obtain finer prototypes. Third, the performance slightly decreases when $n$ is either too small or too large. We argue that: when $n$ is too small, the feature comparison becomes difficult, as background regions may contain diverse semantic information; and when $n$ is too large, the learning task will become too simple, and it will easily fall into over-fitting due to limited semantics on the query images.
\begin{table}[t]
\centering
\caption{Performance~(mIoU\%) comparison with different values of $n$~(the number of background prototypes). $n$ = 0 is equivalent to the baseline method.}
\label{Number of $n$}
\begin{tabular}{@{}c|cccc|c@{}}
\toprule
$n$      & \multicolumn{1}{l}{Fold-0} & \multicolumn{1}{l}{Fold-1} & \multicolumn{1}{l}{Fold-2} & \multicolumn{1}{l|}{Fold-3} & \multicolumn{1}{l}{Mean} \\ \midrule
0      & 61.7                      & 69.5                      & 55.4                      & 56.3                       & 60.8                    \\ 
1      & 61.7                      & 70.1                      & 60.4             & 55.0                       & 61.8                    \\
2      & 61.9                      & 70.0                      & 61.6                      & 55.8                       & 62.3                    \\
3      & 62.2             & \textbf{70.5}             & 61.1                      & \textbf{58.1}              & \textbf{63.0}           \\
4      & 62.0                      & 70.3                      & \textbf{62.0}                      & 56.1                       & 62.6                    \\
5      & 61.3                      & 70.1                      & 60.1                      & 56.6                       & 62.0     \\ \bottomrule
\end{tabular}
\end{table}

\begin{table}[h]
\centering
\caption{Effect of feature alignment scheme on the class-agnostic branch.}
\label{table:source}
\begin{tabular}{@{}c|ccccc@{}}
\toprule
Assignment Scheme & Fold-0 & Fold-1 & Fold-2 & Fold-3 & Mean \\ \midrule
Fig.\ref{fig:compareFeature}~(a)  & 61.7        &  69.5      &  55.4       & 56.3        & 60.8     \\ 
Fig.\ref{fig:compareFeature}~(b)  & 59.8    & 69.7    & 58.2    & 56.7    & 61.1 \\
Fig.\ref{fig:compareFeature}~(c)   & 61.7        &  70.1       &  60.4       & 55.0        & 61.8     \\ 
Fig.\ref{fig:compareFeature}~(d)  & \textbf{62.2}        &  \textbf{70.5}       &  \textbf{61.1}       & \textbf{58.1}        & \textbf{63.0}     \\ 
\bottomrule
\end{tabular}
\end{table}

\begin{table}[]
\caption{Effect of foreground feature alignment scheme on the class-agnostic branch.}
\label{tab:filledPrototype}
\centering
\begin{tabular}{@{}c|cccc|c@{}}
\toprule
Assignment Scheme & Fold-0 & Fold-1 & Fold-2 & Fold-3 & Mean \\ \midrule
Deterministic~(support)   & 61.8   & 69.6   & 59.8   & 57.2   & 62.1 \\
Deterministic~(query)     & 62.2   & 70.0   & 60.1   & 57.4   & 62.4 \\
Random~(query)           & 62.2   & 70.5   & 61.1   & 58.1   & 63.0 \\ \bottomrule
\end{tabular}
\end{table}

\begin{table}[t]
\caption{The influences of hyper-parameter ($\lambda$) for our APANet.}
\label{table:loss-weight}
\centering
\begin{tabular}{@{}c|cccc|c@{}}
\toprule
$\lambda$ & Fold-0         & Fold-1         & Fold-2         & Fold-3         & Mean           \\ \midrule
0      & 61.7          & 69.5          & 55.4          & 56.3          & 60.8          \\
0.1    & 61.1          & 69.9          & 58.2          & 57.3          & 61.6          \\
0.3    & 60.8          & 70.1          & 60.5          & 57.0          & 62.1          \\
0.5    & \textbf{62.2} & \textbf{70.5} & 61.1          & \textbf{58.1} & \textbf{63.0} \\
0.7    & 61.2          & 70.3          & \textbf{61.4} & 57.2          & 62.5          \\
0.9    & 59.8          & 69.2         & 60.0          & 53.3          & 57.7          \\
1.0    & 17.2          & 28.8          & 33.2          & 21.0          & 25.1          \\ \bottomrule
\end{tabular}
\end{table}

\noindent\textbf{Effects of Loss Weight.}
We explore the influence of hyper-parameter $\lambda$, where $\lambda$ is a weight that balances the importance between
two losses. From Table~\ref{table:loss-weight}, we can observe the followings.
First, when $\lambda$ is 0.5, \ie the two losses are equally important, our method yields a 2.2 mIoU improvement over the baseline. This implies the effectiveness of the second loss $L_2$. Second, the performance begins to decrease when $L_2$ gets more attention. Extremely, when we set $\lambda$ to 1, the performance drops rapidly (from 63.0 to 25.1). The reason is that now the whole network learns feature comparison only from the query image without any information from the support images. Third, the performance drops as $\lambda$ decreases. When $\lambda$ is 0, the overall network will degenerate to the baseline that learns feature comparison only on the support-query image pair. 

\begin{figure}[t]
\centering
\centerline{\includegraphics[width=1.02\linewidth]{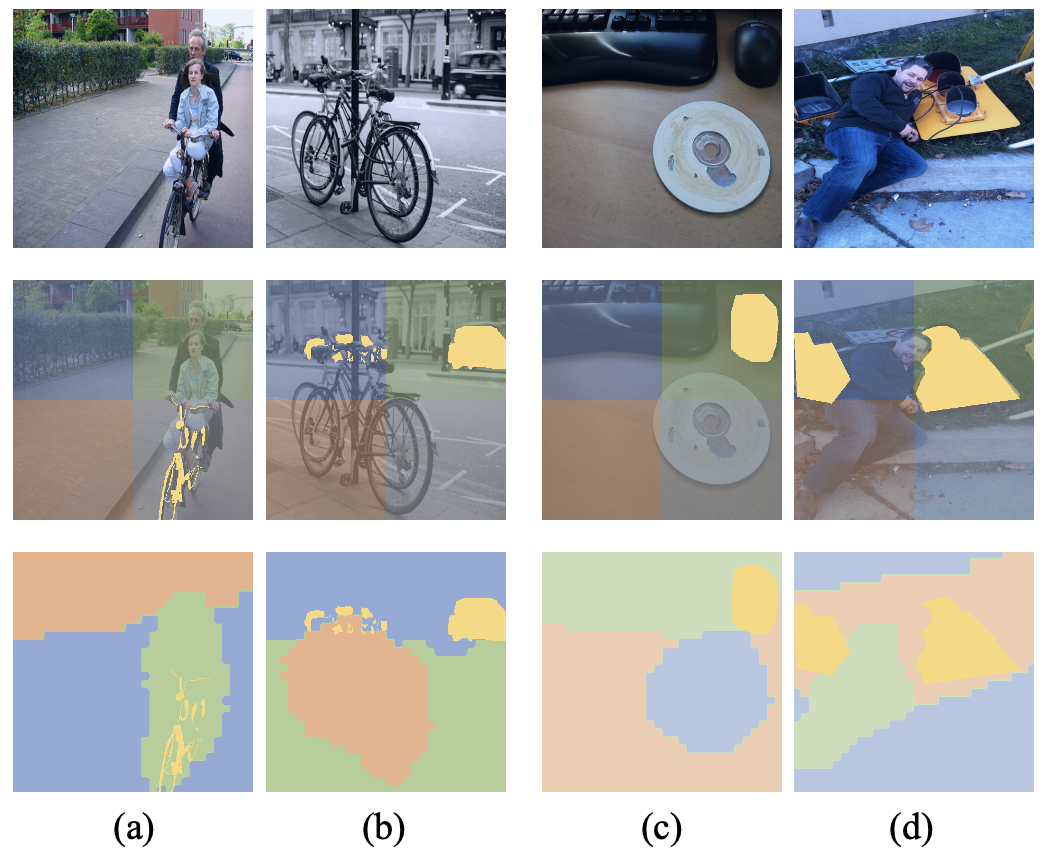}}
\caption{Visualization of two types of class-agnostic mask generation methods. The selected images are shown in the first row, the first two columns are from PASCAL-5$^i$ and the last two columns are from COCO-20$^i$. Note that the foreground objects in the second and third rows are covered by yellow masks (ground-truth). The second row shows multiple class-agnostic masks generated by the SPP method (2$\times$2 spatial size). The third row shows the visualization of the $k$-means clustering ($k$=3) with high-level features. We can see that each clustering region may correspond to a specific semantic when we use the clustering method, which is more reasonable than the simple SPP.}
\label{fig:KM_graph}
\end{figure}

\noindent\textbf{Class-agnostic Prototypes Alignment Methods.}
\begin{table}[t]
\centering
\caption{Results using spatial pyramid pooling prototypes.}
\label{table:spp}
\begin{tabular}{@{}c|cccc|c@{}}
\toprule
Spatial Size    & Fold-0 & Fold-1 & Fold-2 & Fold-3 & Mean \\ \midrule
2$\times$2 & 60.1    & 69.1    & 57.4    & 54.0    & 60.1 \\
3$\times$3 & 60.6    & 69.2    & 60.2    & 55.6    & 61.4 \\
4$\times$4 & \textbf{60.9}    & \textbf{70.0}    & \textbf{60.7}    & 56.1    & \textbf{61.9} \\
5$\times$5 & 59.8        &  69.7       &  58.2       & \textbf{56.7}        & 61.1     \\ \bottomrule
\end{tabular}
\end{table}
Instead of extracting the class-agnostic prototypes from the support features~\cite{panet, PL, fwb, simpropnet}, we generate the class-agnostic prototypes from the query features to completely learn the feature comparison. Which strategy is better? To answer this question, we conduct experiments based on the support/query class-agnostic prototypes. The number of clusters is set to 1 in our APANet to eliminate the influence of clustering. Table~\ref{table:source} reports the results of these two strategies. It shows that using the class-agnostic prototype from the query features outperforms its counterpart. This is because it is not appropriate to derive the background prototype for the query image from the support image as we discussed. 

On the other hand, negative pairs are constructed by assigning suitable prototypes for the foreground features of the query image. To this end, we randomly select one class-agnostic prototype and densely pair it with each location of foreground features of the query image. To verify the effectiveness of random assignment strategy, we re-design a deterministic assignment scheme. We first mask average pool~(MAP) the whole support or query background features and derive a deterministic background prototype; then, we assign the deterministic background prototype to each foreground query feature. Here, we fix the positive pairs construction method and set the number of clusters to 3. The experimental results are reported in Table~\ref{tab:filledPrototype}. We can see that the manner of random assignment achieves the best performance. Compared with deterministic assignment, random class-agnostic prototype can represent the background with multiple semantic clusters and thus can construct more diverse sample pairs to train an unbiased classifier. In addition, there are similar results when we use the deterministic prototypes from query and support background features (62.4 vs. 62.1). Note that the same category object always appears in the support and query images, and it is reasonable to construct a negative sample for each foreground feature no matter which prototype (support or query background prototype) is assigned.

\noindent\textbf{Importance of Prototype Generation Method.}
In our method, the class-agnostic prototypes are generated by using the $k$-means clustering algorithm on the query feature map. We conduct another generation approach to see whether the performance is sensitive to the generation strategy. Different from the clustering algorithm that adapts the assignments according to spatial semantic distribution, we simply apply spatial pyramid pooling~(SPP)~\cite{he2015spatial} on a query feature map to obtain class-agnostic prototypes. The visualization of two types of class-agnostic mask generation methods is shown in Fig.~\ref{fig:KM_graph}. With a pyramid level of $a\times a$ bins, we implement the average pooling at each bin and obtain a total of $a^2$ prototypes. In our experiment, we use 4-level pyramids: \{2$\times$2, 3$\times$3, 4$\times$4, 5$\times$5\}. 

In Table~\ref{table:spp}, we can see that using the spatial class-agnostic prototypes generation is comparable to the baseline (61.9\% vs. 60.8\%). This further demonstrates the effectiveness of learning a complete feature comparison. In addition, our method is still better than spatial assignment under all spatial sizes (63.0\% vs. 61.9\%), which implies that it is better to use a semantic clustering strategy for class-agnostic prototype generation.

\section{Conclusion}  
In this paper, we embed complementary feature comparison into the metric-based few-shot semantic segmentation (FSS) framework to improve the FSS performance.
Specifically, unlike previous works unilaterally predicting the foreground mask with the prototypes extracted for foreground objects only, we propose to compute class-agnostic prototypes and construct complementary sample pairs, which enables us to perform a complementary feature comparison with a two-branch network architecture, \ie, the class-specific and class-agnostic branches. To ensure the similarity between the prototype and the background features during prediction, we propose to extract class-agnostic prototypes from the query features alone and then conduct feature comparison in a self-contrastive manner. The proposed network APANet achieves the state-of-the-art performance on both the PASCAL-5$^i$ and COCO-20$^i$ datasets, which validates the effectiveness of our method.

\ifCLASSOPTIONcaptionsoff
  \newpage
\fi

\bibliographystyle{IEEEtran}
\bibliography{egbib}

\end{document}